\documentclass[conference]{IEEEtran}
\IEEEoverridecommandlockouts
\usepackage{cite}
\usepackage{amsmath,amssymb,amsfonts}

\usepackage{textcomp}
\usepackage{xcolor}
\usepackage{graphicx}
\usepackage{url}
\usepackage{subcaption} 
\usepackage{lipsum}  
\usepackage{algorithm}
\usepackage[noend]{algpseudocode}
\usepackage{booktabs}
\usepackage{hyperref}

\def\BibTeX{{\rm B\kern-.05em{\sc i\kern-.025em b}\kern-.08em
    T\kern-.1667em\lower.7ex\hbox{E}\kern-.125emX}}
\begin{document}

\title{Dynamic Reward Scaling for Multivariate Time Series Anomaly Detection: A VAE-Enhanced Reinforcement Learning Approach}

\author{

\IEEEauthorblockN{1\textsuperscript{st} Bahareh Golchin}
\IEEEauthorblockA{\textit{School of Computer Science} \\
\textit{Portland State University}\\
Portland, Oregon, United States \\
bgolchin@pdx.edu}
\and
\IEEEauthorblockN{2\textsuperscript{nd} Banafsheh Rekabdar}
\IEEEauthorblockA{\textit{School of Computer Science} \\
\textit{Portland State University}\\
Portland, Oregon, United States \\
rekabdar@pdx.edu}

}

\maketitle

\begin{abstract}
\label{sec:abstract}
Detecting anomalies in multivariate time series is essential for monitoring complex industrial systems, where high dimensionality, limited labeled data, and subtle dependencies between sensors cause significant challenges. This paper presents a deep reinforcement learning framework that combines a Variational Autoencoder (VAE), an LSTM-based Deep Q-Network (DQN), dynamic reward shaping, and an active learning module to address these issues in a unified learning framework. The main contribution is the implementation of Dynamic Reward Scaling for Multivariate Time
Series Anomaly Detection (DRSMT), which demonstrates how each component enhances the detection process. The VAE captures compact latent representations and reduces noise. The DQN enables adaptive, sequential anomaly classification, and the dynamic reward shaping balances exploration and exploitation during training by adjusting the importance of reconstruction and classification signals. In addition, active learning identifies the most uncertain samples for labeling, reducing the need for extensive manual supervision. Experiments on two multivariate benchmarks, namely Server Machine Dataset (SMD) and Water Distribution Testbed (WADI), show that the proposed method outperforms existing baselines in F1-score and AU-PR. These results highlight the effectiveness of combining generative modeling, reinforcement learning, and selective supervision for accurate and scalable anomaly detection in real-world multivariate systems.\footnote{Code is available at
\href{https://github.com/baharehgl/Dynamic-Reward-RL-VAE}{GitHub Repository}.}
\end{abstract}


\section{Introduction}
\label{sec1:Introduction}

In many of today's applications, identifying and removing anomalies (i.e., outliers) has become essential to ensure system reliability. In multivariate time series data, specifically, different factors can result in anomalies. These factors include equipment malfunction, sensor failure, and errors made by operators \cite{alzarooni2025review}. Over the years, several machine learning algorithms have become well-suited for detecting anomalies. 

Detecting outliers in multivariate time series data has many practical applications. Some examples include 1) monitoring the status of various equipment, 2) spotting irregularities in Internet of Things sensor networks, 3) observing patients' health metrics, and 4) identifying cyber risks in complex systems. Multivariate time series anomaly detection (MTSAD) has been the focus of many studies in the literature. Many different approaches have been introduced to address the difficulty of finding rare and unexpected events in complex and noisy sensor data streams \cite{abshari2025surveycps}.

Traditionally, statistical methods and algorithms have been developed to identify abnormal behavior in individual sensors \cite{darban2022deeplearningtsad}. However, recent progresses in machine learning techniques have been proven to be able to tackle different multivariate anomaly detection problems very effectively \cite{adewuyi2024precision}. In specific, for multivariate time series with complex nonlinear temporal dynamics and interconnected sensor relationships, deep learning methods have been remarkably effective \cite{zheng2024gstpro}. Most of these models first learn the normal behavior of data from unlabeled datasets. Then, the model can potentially find samples that deviate from normal behavior as anomalies \cite{bouman2024unsupervised} and \cite{nguyen2024vaead}.

In real-world scenarios, specifically, in multivariate time series, there is often a lack of labeled data. This issue impairs the model to distinguish between normal and abnormal behaviour. In industrial environments with dozens of synchronized sensor channels, the normal boundary is often unclear across multiple dimensions. This could potentially lead to even small deviations being classified as anomalies. Models that cannot distinguish between normal and anomaly classes could potentially predict normal samples as anomalous. This will result in high false positives. This challenge is particularly severe in multivariate settings where anomalies often take place through unexpected combinations of sensor values rather than individual sensor deviations \cite{rebjock2021online}.

Current multivariate anomaly detection models, which are proposed in the literature, face additional challenges, including computational complexity and the curse of dimensionality. As the number of sensor features increases, data sparsity increases. This makes it harder to distinguish genuine anomalies from normal variations. Furthermore, in complex industrial systems like water treatment plants and server monitoring environments, anomalies take place frequently. This poses significant challenges due to the anomalies being interconnected as well as devices with many parameters having complex temporal correlations \cite{birihanu2024explainable}.

Our proposed algorithm tackles the aforementioned challenges in detecting anomalies in multivariate datasets through a novel integration of three key components: 1) a Variational Autoencoder (VAE), 2) an LSTM-based Deep Q-Network (DQN), and 3) an active learning mechanism with dynamic reward scaling. The algorithm is performed by first training a VAE on normal multivariate sensor data to learn the underlying patterns and relationships between multiple sensor channels. During the anomaly detection phase, the system uses sliding windows of multivariate time series data as states. In this phase, each window contains synchronized readings from all sensors. The DQN, powered by LSTM layers to capture temporal dependencies, makes binary classification decisions (normal or anomalous) for each time step. The core innovation is in the dynamic reward scaling mechanism, where the total reward combines classification accuracy rewards with VAE reconstruction error penalties. This is scaled by an adaptive coefficient $\lambda(t)$ that automatically balances exploration of uncertain patterns with exploitation of learned knowledge.

We specifically address the major challenges of multivariate time series anomaly detection through several key mechanisms. First, to handle the curse of dimensionality common in high dimensional sensor data, the VAE component learns a compact latent representation. This captures the essential multivariate relationships while reducing computational complexity \cite{wang2025survey} and \cite{iqbal2024deepensemble}. 

Second, the dynamic reward scaling addresses the challenge of sparse anomalies in multivariate settings by automatically adjusting the influence of reconstruction error during training. In this phase, the algorithm initially emphasizes on the exploration of normal patterns across all sensor channels, then gradually focuses on precise anomaly classification as the agent learns \cite{cheng2025inverse}.

Third, the active learning component tackles the labeling challenge in multivariate systems by intelligently selecting the most uncertain multivariate patterns for human annotation. This helps reduce the amount of labeled data needed as well as maintain high detection accuracy across multiple sensor dimensions \cite{golchin2024anomaly}. 

Our proposed method (DRSMT) helps the system to identify hard-to-detect anomalies that manifest through unexpected combinations of sensor values rather than individual sensor deviations, which is particularly important in industrial environments like water treatment plants and server monitoring systems \cite{mozaffari2022online}.

In what follows, we summarize the main contributions of our proposed method.
\begin{itemize}
\item \textbf{Novel Dynamic Reward Scaling Framework for Multivariate Anomaly Detection}. We developed the first dynamic reward scaling mechanism that automatically balances exploration and exploitation in multivariate time series anomaly detection through an adaptive coefficient $\lambda(t)$ that adjusts the influence of VAE reconstruction error during training.

\item \textbf{Efficient Integration of VAE and Reinforcement Learning for High Dimensional Sensor Data}. We successfully implemented a VAE to receive multivariate time series data as input and produce extra feedback to use it into DQN as a reward.

\item \textbf{Active Learning Strategy for Minimal Labeling in Multivariate Industrial Systems}. We developed an uncertainty-based active learning approach specifically designed for multivariate anomaly detection that uses margin-based sampling to identify the most informative multivariate patterns.
\end{itemize}

The following sections are arranged as follows. In Section \ref{sec2:related}, we provide an overview of the literature related to our work. Moreover, the background of time series anomaly detection is reviewed in Section \ref{sec3:background}. Next, we present our proposed framework in details in Section \ref{sec4:ProposedMethod}. Section \ref{sec5:Experiment} examines the implementation details of our proposed method. Finally, we conclude our study in Section \ref{sec6:Conclusion}.

\section{Related Work}
\label{sec2:related}
Our study falls within the intersection of the following four bodies of literature.

\subsection{Statistical and Traditional Machine Learning Approaches}
Traditionally, this problem has depended heavily on statistical methods and classical machine learning approaches to identify unusual patterns in data. Statistical-based methods build models from given datasets and apply statistical tests to determine whether unseen data conforms to the proposed model. A fundamental assumption in these models is that they presume the normal data follows a specific probability distribution, including parametric models. Some examples are Gaussian models, regression models, and logistic regression \cite{golchin2024anomaly} and \cite{sanami2024calibrated}. However, an important limitation of statistical approaches is the basic assumption that normal behavior follows an existing distribution. This often does not hold in complex, real-world scenarios\cite{park2020lstm}.

Machine learning approaches address these limitations by using labeled training data to differentiate between normal and abnormal instances through classification or clustering algorithms. Some examples include Bayesian networks, rule-based systems, support vector machines, and neural networks\cite{wei2024anomaly} and \cite{su2023combining}. 

\subsection{Deep Learning Methods for Time Series Anomaly Detection}Recent progress in deep learning techniques have effectively tackled various anomaly detection problems, particularly for time series with complex nonlinear temporal dynamics \cite{wang2025survey}. Deep learning techniques focus on understanding typical patterns in data without using labeled data. Therefore, they may detect samples that do not follow normal patterns and mark them as anomalies \cite{zamanzadeh2024carla}. Techniques like autoencoders, VAEs, recurrent neural networks (RNNs), LSTM networks, generative adversarial networks (GANs), and transformers have successfully demonstrated the capacity to detect anomalies in time series datasets \cite{niu2020lstmvaegan} and \cite{jeong2023anomalybert}.

In the literature, deep models like LSTM-VAE are preferred. The reason for it is that they have been excellent at minimizing forecasting errors, and at the same time, capturing temporal dependencies in time series data \cite{niu2020lstmvaegan}. These models can be classified into reconstruction-based approaches that learn to recreate normal patterns and forecasting-based methods that predict future values to identify deviations \cite{wang2025survey}.

\subsection{Multivariate Time Series Anomaly Detection Challenges}
Multivariate time series anomaly detection presents unique challenges compared to univariate approaches because it requires understanding both temporal dependencies within individual sensors and spatial relationships between multiple variables \cite{wei2024anomaly} and \cite{ahmad2017unsupervised}. In multivariate systems with dozens of synchronized sensor channels, the boundary for normal behavior is often narrowly defined across multiple dimensions\cite{pang2021deep}. In this case, even small deviations might be wrongly identified as anomalies. Current multivariate anomaly detection methods face additional challenges. Some of these challenges include computational complexity and the curse of dimensionality, where anomalies often manifest through unexpected combinations of sensor values rather than individual sensor deviations \cite{su2019robust} and \cite{tuli2022tranad}. Recent studies show that GNNs are effective in learning relationships between variables in multivariate time series. Some of the methods include Graph Deviation Network learning graph structures representing relationships between channels \cite{lu2023multiscale}.

\subsection{RL and Active Learning Integration}
The integration of RL with active learning represents recent progress in time series anomaly detection, which addresses the challenge of limited labeled data in real-world scenarios\cite{wu2021rlad}. Deep Reinforcement Learning (DRL) models can effectively use a limited set of labeled anomalous data while extensively exploring large pools of unlabeled data. This enables the detection of new anomaly types not present in the labeled dataset and \cite{wu2023timesnet}. Methods like RLAD, with the combination of DRL and active learning, efficiently identify and respond to anomalies \cite{wu2021rlad}.

\section{Background}
\label{sec3:background}

This section provides essential theoretical foundations for our multivariate time series anomaly detection framework, which integrates DQN, VAE, dynamic reward scaling, and active learning for industrial sensor data.

\subsection{DQNs for Sequential Decision Making}

In this study, we treat the problem of multivariate anomaly detection using a Markovian Decision Process Framework where an agent observes sliding windows of synchronized sensor data and classifies each time step as normal or anomalous.  The agent learns an action‐value function \(Q(s, a)\) that estimates expected rewards for taking action \(a\) in state \(s\).  The Q‐function is updated using the Bellman equation:
\[
Q^*(s,a) \;=\; \mathbb{E}\bigl[r(s,a) + \gamma \max_{a'} Q^*(s',a')\bigr],
\]
where \(r(s,a)\) is the immediate reward and \(\gamma\) is the discount factor.  To handle the complexity of multivariate sensor data, we implement the Q‐network using LSTM layers that capture temporal dependencies across sensor channels.  DQN employs two key stabilization techniques: experience replay using transition tuples \(\langle s, a, r, s'\rangle\) and a target network that provides fixed reference values during training updates.

\subsection{VAE for Multivariate Reconstruction}

VAEs learn compact latent representations of normal multivariate sensor patterns through encoder–decoder architectures.  The encoder maps input windows \(x\) to latent distributions characterized by mean \(\mu(x)\) and variance \(\sigma^2(x)\), while the decoder reconstructs the original input.  The VAE optimizes the Evidence Lower Bound (ELBO):
\[
\mathcal{L}(\theta,\phi;x)
=\;\mathbb{E}_{q(z\mid x;\phi)}\bigl[\log p(x\mid z;\theta)\bigr]
\;-\;\mathrm{KL}\bigl[q(z\mid x;\phi)\,\|\,p(z)\bigr].
\]
For anomaly detection, the reconstruction error
\(\|x - \widehat{x}\|^2\) serves as an unsupervised anomaly score: normal patterns yield low error, while anomalies produce high error, guiding the RL agent.

\subsection{Dynamic Reward Scaling Mechanism}
Reward shaping is a method used in RL to help the agent learning by changing the reward system to include domain knowledge. The main goal of reward shaping is speeding up learning and improving performance through intermediate reward signals that encourage good behaviors, especially when final reward signals are rare or delayed. For example, potential-based reward shaping (PBRS) ensures the optimal policy stays unchanged while adding rewards to support specific paths. This method has become widely accepted in RL tasks where environmental basic rewards are insufficient for effectively guiding agent exploration \cite{hong2025adaptive}. 

\subsection{Active Learning for Minimal Supervision}

Active learning improves machine learning efficiency by selectively including unlabeled data for manual labeling. Given a labeled dataset \( L = (X, Y) \) and an unlabeled pool \( U = (x_1, x_2, \dots, x_n) \), active learning uses a query function \( Q \) to identify the most informative samples from \( U \). These selected instances are labeled by experts and are incorporated into the training set, which refines the classifier \( C \) with minimal labeled data. One of the key query strategies is Margin Sampling. In this strategy, samples are chosen with the smallest confidence gap between the top two predicted classes: $x_m = \arg \min (P_C(\hat{y}_1 \mid x) - P_C(\hat{y}_2 \mid x))$. Querying strategies ensure efficient model training.

\section{Proposed Method}
\label{sec4:ProposedMethod}
\begin{figure*}[ht]
    \centering
    \includegraphics[width=0.8\textwidth]{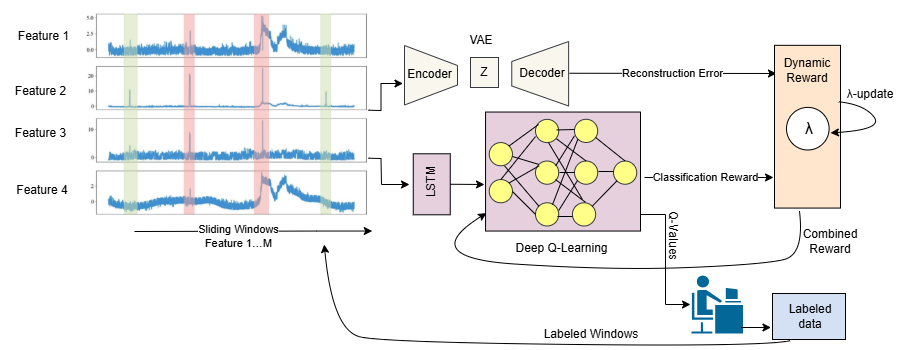}
      \caption{Workflow of our proposed method (DRSMT). A multivariate \(N_{\mathrm{steps}}\times M\) sliding window (with \(M\) sensor channels) is fed in parallel to: 1) a VAE that produces a reconstruction error penalty \(R_{2}\), and 2) an LSTM‐based DQN that outputs classification rewards \(R_{1}\). The Dynamic Reward module combines \(R_{1}\) and \(R_{2}\) with an adaptive coefficient \(\lambda(t)\), which is automatically updated during training to trade off exploration (novelty via reconstruction error) and exploitation (correct classification). An Active Learning loop queries the most uncertain windows for human labeling, closing the loop with minimal labeled data.}
     \label{figure 1}
\end{figure*}

In this section, we describe our proposed method in detail. Our framework extends the literature on the univariate dynamic reward scaling approach to handle multivariate time series data by combining a VAE, DRL, active learning, and dynamic reward shaping. Figure \ref{figure 1} depicts the workflow of our proposed multivariate anomaly detection system.

\subsection{ Implementing Multivariate Anomaly Detection with VAE}
For multivariate time series data, each input x represents a sliding window of length n steps containing synchronized readings from multiple sensor channels. Unlike the univariate case, our multivariate approach processes windows of shape $(n, d)$, where $d$ represents the number of sensor features. The VAE is trained exclusively on normal multivariate patterns to learn a compact latent representation that captures both temporal dependencies within individual sensors and spatial relationships between different sensor channels.

During preprocessing, we apply feature selection to remove sensors with zero variance across the training samples to avoid problems that come with too many dimensions. Each sliding window is flattened into a one-dimensional vector of size $(n \times d)$ before being fed to the VAE. 

The VAE architecture consists of an encoder that maps the flattened multivariate window to latent distributions characterized by mean $\mu(x)$ and variance $\sigma^2(x)$, and a decoder that reconstructs the original input. The reconstruction error serves as an unsupervised anomaly score, where normal multivariate patterns exhibit low reconstruction error while anomalous combinations of sensor values produce high reconstruction error. This reconstruction error is integrated into the RL agent's reward function to guide the learning process toward accurate anomaly classification in the multivariate space.

\subsection{Implementing Deep RL with Dynamic Reward Shaping for Multivariate Anomaly Detection} \label{DRL-Implementation}

Our approach formulates multivariate anomaly detection as a sequential decision‐making task where a Deep RL agent observes synchronized sensor data from multiple channels and classifies each time step as normal or anomalous.  The agent receives reward signals that combine classification accuracy with VAE‐based reconstruction penalties, dynamically scaled to balance exploration and exploitation throughout training.

\textbf{State Representation.} Each state \(s_t\) corresponds to a sliding window of length \(N_{\text{STEPS}}\) from the multivariate time series.  For a system with \(d\) sensors, the state contains synchronized readings:
\[
s_t = \{\,x_{t - N_{\text{STEPS}} + 1}, \ldots, x_t\},\quad x_i\in\mathbb{R}^d.
\]
To allow the agent to distinguish its prediction action, we augment each state with an action indicator:
\[
s^0_t = [\,s_t;\,\mathbf{0}\!],\quad
s^1_t = [\,s_t;\,\mathbf{1}\!],
\]
so that \(s^a_t\in\mathbb{R}^{N_{\text{STEPS}}\times(d+1)}\).

\textbf{Action Space.} At each time \(t\), the agent takes a binary action $
a_t \;\in\;\{0,1\}$, where $0$ represents predict normal, and $1$ represents predict anomalous. After each action, the environment shifts the sliding window forward.

\textbf{Policy and Q‐Network.} The policy \(\pi(a\mid s)\) is derived from an LSTM‐based Q-network that captures temporal dependencies.  The network processes the augmented state through LSTM layers (64 hidden units), then a dense layer outputs $Q^\pi(s_t,0),\; Q^\pi(s_t,1)$. Actions are selected via \(\varepsilon\)-greedy exploration:
\[
a_t = 
\begin{cases}
\arg\max_{a\in\{0,1\}}Q^\pi(s_t,a), & \text{with probability }1-\varepsilon,\\
\text{random }\{0,1\},             & \text{with probability }\varepsilon.
\end{cases}
\]

\textbf{Dynamic Reward Mechanism.} We combine immediate classification rewards (i.e., extrinsic reward, $R_1$) with VAE reconstruction penalties (i.e., intrinsic reward, $R_2$).  The classification reward is defined piecewise as:

\begin{equation}
R_1(s_t, a_t) =
\begin{cases} 
TP_{val} & \text{if } a_t = 1 \text{ and } y_t = 1, \\ 
TN_{val} & \text{if } a_t = 0 \text{ and } y_t = 0, \\ 
FP_{val} & \text{if } a_t = 1 \text{ and } y_t = 0, \\ 
FN_{val} & \text{if } a_t = 0 \text{ and } y_t = 1.
\end{cases}
\end{equation}
where, 1) $TP_{val} = 10, TN_{val} = 1$, and 2) $FP_{val} = -1, FN_{val} = -10$. This implementation creates a reward vector where index 0 represents the reward for non-anomaly classification and index 1 for anomaly classification. The asymmetric reward structure reflects the higher importance of detecting true anomalies.

Next, Reconstruction-based reward shaping in our proposed method uses the VAE to guide the learning process by incorporating reconstruction error as an additional reward component. The VAE is trained on normal time series data to learn a compact latent representation. This enables the VAE to reconstruct normal patterns effectively. The reconstruction error is calculated as the MSE between the original input $x_t$ and its reconstruction $\hat{x}_t$, which serves as a measure of how well the current state aligns with normal behavior. The reconstruction error is computed as:
\begin{equation}
   R_2(s_t, a_t)=\text{MSE}(x_t, \hat{x}_t) = \frac{1}{n} \sum_{i=1}^{n} (x_{t,i} - \hat{x}_{t,i})^2, 
\end{equation}
where $n$ is the dimensionality of the input window.

Finally, the mathematical formulation for the total reward is:
\begin{equation}
    R_{total}(s_t,a_t)= R_1(s_t,a_t)+\lambda(t)R_2 (s_t,a_t)
\end{equation}
where $\lambda(t)$ is a dynamic scaling coefficient that adjusts the effect of the reconstruction penalty over time. In our proposed method, the dynamic coefficient $\lambda(t)$ plays a crucial role in balancing the influence of the reconstruction error from VAE in the total reward calculation. 

Reconstruction error provides an unsupervised signal that complements the supervised classification reward $R_1(s_t,a_t)$. Without scaling by $\lambda(t)$,  the magnitude of the reconstruction error might dominate or be negligible compared to $R_1(s_t,a_t)$, which could lead to suboptimal learning.

By dynamically adjusting $\lambda(t)$, the framework ensures that: 1) The agent explores normal patterns early in training, and 2) the focus gradually shifts toward accurate anomaly classification as training progresses.

The coefficient (i.e., $\lambda(t)$) is updated after each episode based on the total episode reward. The update rule follows a proportional control mechanism:

\begin{equation}
\lambda_{t+1} = \text{clip}\left(\lambda_t + \alpha \left(R_{\text{target}} - R_{\text{episode}}\right), \lambda_{\text{min}}, \lambda_{\text{max}}\right),
\end{equation}

where: 1) $R_{\text{target}}$: Target reward for an episode. 2) $R_{\text{episode}}$: Total reward achieved in the current episode. 3) $\alpha$: Learning rate for adjusting $\lambda(t)$. 4) $\lambda_{\text{min}}$ and $\lambda_{\text{max}}$: Minimum and maximum allowable values for $\lambda(t)$.

This formula ensures that:
\begin{itemize}
    \item If $R_{\text{episode}} < R_{\text{target}}$, then $\lambda(t)$ increases to emphasize reconstruction error.
    \item If $R_{\text{episode}} > R_{\text{target}}$, then $\lambda(t)$ decreases to reduce reliance on reconstruction error.
\end{itemize}

\begin{figure}[htbp]
    \centering
    \begin{subfigure}[b]{0.45\textwidth}
        \centering
        \includegraphics[width=\textwidth]{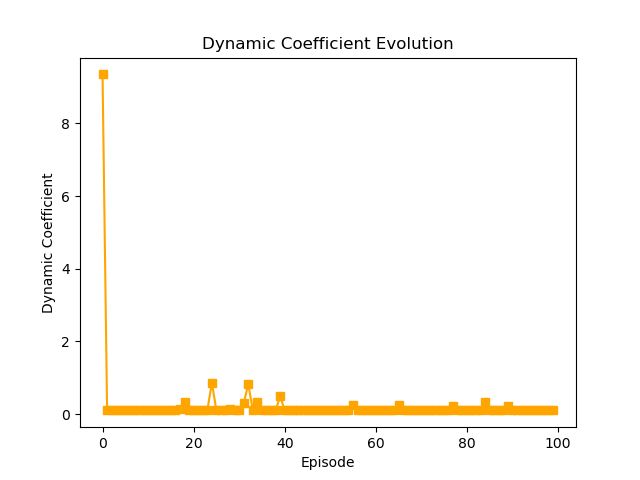}
        \caption{Dynamic coefficient evolution over episodes}
        \label{fig:sub1-coef-reward curve}
    \end{subfigure}
    \hfill
    \begin{subfigure}[b]{0.45\textwidth}
        \centering
        \includegraphics[width=\textwidth]{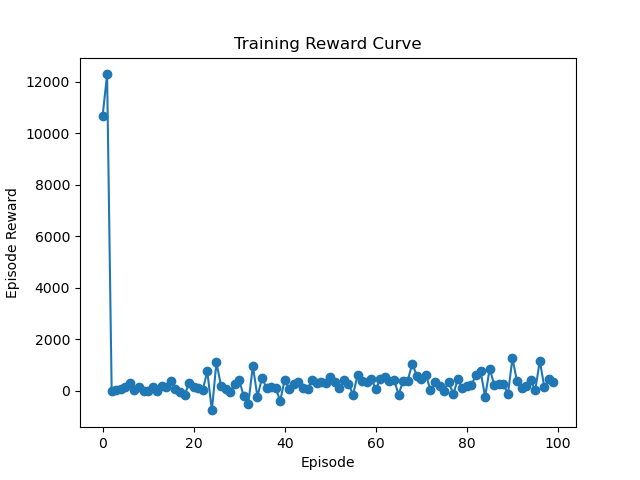}
        \caption{Training reward curve}
        \label{fig:sub2-coef-reward curve}
    \end{subfigure}
    \caption{Relationship between the dynamic coefficient and reward evolution during training.}
    \label{fig:coef-reward curve}
\end{figure}

Figure \ref{fig:coef-reward curve} depicts the relationship between the dynamic coefficient evolution and the training reward during RL training. Figure \ref{fig:sub1-coef-reward curve} shows how the scaling factor $\lambda(t)$ decreases over episodes. It starts at a high value to prioritize exploration (via reconstruction error) and gradually stabilizes as the agent shifts focus to exploitation (classification accuracy). This behavior directly affects figure \ref{fig:sub2-coef-reward curve}, where initial episodes show higher rewards due to the significant contribution of reconstruction error ($R_2$) scaled by $\lambda(t)$. As $\lambda(t)$ decreases, the reward curve stabilizes and reflects classification performance ($R_1$), with fluctuations arising from variations in state-action transitions. These figures demonstrate how the dynamic reward mechanism effectively balances exploration and exploitation throughout training.

\subsection{Implementing Active Learning for Anomaly Detection}
In our proposed method, the active learning module is designed to iteratively identify and label the most uncertain samples from the time series dataset to improve the agent's anomaly detection capabilities. The active learning class uses a margin-based sampling strategy. It calculates the absolute difference between the Q-values of the two possible actions (normal or anomaly) for each state (i.e., $\text{Margin}(s) = |Q(s, a_1) - Q(s, a_2)|$). First, samples with the smallest margin (i.e., the most uncertain predictions) are ranked. Then, the top-N uncertain samples are selected for manual labeling by a user (i.e., $\text{Selected Samples} = \arg \min_{s \in \mathcal{S}} \text{Margin}(s)$). This process ensures that the agent focuses on learning from the most ambiguous cases. This accelerates the agent's capability in detecting anomalies.

Once these samples are labeled, they are added back to the dataset. Then, label propagation is applied using a semi-supervised learning technique (i.e., LabelSpreading) to transmit labels to nearby unlabeled samples based on feature similarity. The probability of a label $y_i$ for an unlabeled sample $x_i$ is computed as:

\begin{equation}
   P(y_i | x_i) = \frac{\sum_{j \in \mathcal{L}} w_{ij} P(y_j | x_j)}{\sum_{j} w_{ij}} 
\end{equation}
 where 1) $\mathcal{L}$ is the set of labeled samples, and 2) $w_{ij}$ is the similarity weight between samples $x_i$ and $x_j$. This combination of active learning and label propagation: 1) reduces reliance on large amounts of labeled data, and 2) enables the agent to learn effectively. The active learning process is tightly integrated into the RL loop, which allows the agent to refine its policy progressively by incorporating high-value labeled samples into its training set. Our proposed method algorithm is detailed in Algorithm \ref{alg:rl-anomaly}.

\begin{figure}[htbp]
    \centering
    \includegraphics[width=0.45\textwidth]{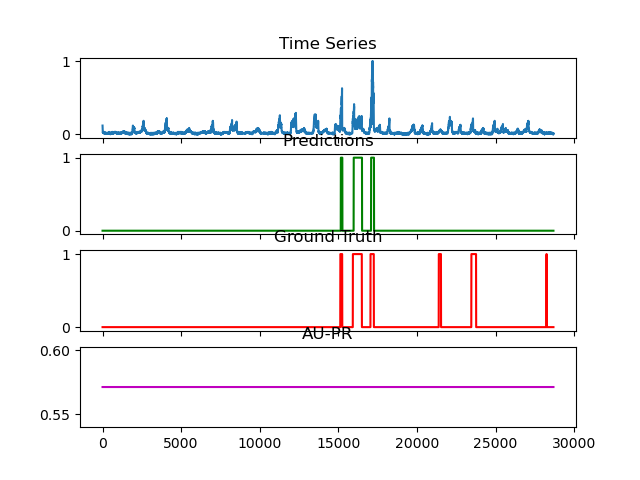}
    \caption{Example visualizations of SMD anomaly detection results from the proposed method.}
    \label{fig:subfigureExample}
\end{figure}

Figure \ref{fig:subfigureExample} illustrates one validation episode on the SMD dataset as an example. In the top panel, the normalized sensor reading over time, with a few sharp spikes corresponding to ground-truth anomalies (third panel, red). In the second panel (green), the binary predictions of our RL policy closely track those true spikes, correcting the anomaly windows and remaining silent elsewhere. The bottom panel plots the area of the episode under the precision-recall curve (AU-PR), which shows strong overall detection performance despite class imbalance. Overall, these plots show that our approach finds each anomaly correctly while keeping false alarms very low throughout the whole sequence.

\begin{algorithm}[htbp]
  \small
  \resizebox{\columnwidth}{!}{%
    \begin{minipage}{\columnwidth}
      \caption{DRSMT: Dynamic Reward Scaling RL with VAE and Active Learning for Multivariate Anomaly Detection}
      \label{alg:rl-anomaly}
      \begin{algorithmic}[1]
        \Require Multivariate time series dataset path $D$, label path $L$, sliding‐window length $n_{\text{steps}}$, RL episodes $N$, batch size $B$, initial dynamic coefficient $\lambda_0$, AL budget $K_{\text{AL}}$, pseudo‐label budget $K_{\text{LP}}$, CV folds $K$
        \Ensure Trained Q‐network; Precision, Recall, F1, AUPR curves

        \Function{BuildVAE}{$D, n_{\text{steps}}$}
          \State Load time series $X$ and labels $y$ from $D,L$
          \State Slide length‐$n_{\text{steps}}$ windows over normal segments ($y=0$)
          \State Standardize and flatten each window $\to$ $\{w_i\}$
          \State Train VAE$(w)$ to minimize reconstruction + KL loss
          \State \Return trained VAE, encoder, feature list
        \EndFunction

        \Function{ComputePenalty}{$\text{VAE},D, n_{\text{steps}}$}
          \State Slide windows over entire $D$, standardize with same scaler
          \State For each batch, compute $r_i = \|w - \widehat w\|^2$
          \State Build penalty array $p[t] = r_{t}$ (zero‐pad first $n_{\text{steps}}-1$)
          \State \Return penalty array $p$
        \EndFunction

        \Function{WarmUp}{$env, M$}
          \State Collect initial states $\{s\}$ by one‐step policy (e.g. IsolationForest outliers)
          \State Play random actions to fill $\mathit{ReplayMem}$ to size $M$
          \State \Return $\mathit{ReplayMem}$
        \EndFunction

        \Function{TrainRL}{$env, \text{VAE}, p, \lambda, N, B, K_{\text{AL}}, K_{\text{LP}}$}
          \State Initialize Q‐network $Q$ and target network $Q'$   
          \For{$e = 1$ \textbf{to} $N$}
            \State \# Active‐Learning + Pseudo‐label
            \State Compute uncertainty on unlabelled states via LabelSpreading 
            \State Label top $K_{\text{AL}}$ points with ground‐truth; next $K_{\text{LP}}$ with LP pseudo‐labels
            \State total\_reward $\gets 0$
            \State Reset $env$; $s \gets env.\text{reset}()$
            \While{not done}
              \State With $\epsilon$‐greedy from $Q(s)$ choose action $a$
              \State Observe next states $\{s_0',s_1'\}$ and labels $y$ in env
              \State Compute extrinsic $r_{\text{cls}}$ from $y$ and scaled penalty $r_{\text{vae}}=\lambda\,p[t]$
              \State $r\gets [r_{\text{cls},0}+r_{\text{vae}},\,r_{\text{cls},1}+r_{\text{vae}}]$
              \State Store $(s,r,s'_a,\text{done})$ in $\mathit{ReplayMem}$
              \State Sample minibatch of size $B$, update $Q$ via Bellman $Q\!-\!\text{MSE}$
              \If{step $\bmod$ $C=0$} \State Sync $Q'\gets Q$
              \EndIf
              \State $s\gets s'_a$, total\_reward $\mathrel{+}=r[a]$
            \EndWhile
            \State $\lambda \gets \text{clip}\bigl(\lambda + \alpha\,(R_{\text{target}} - \text{total\_reward}),\,\lambda_{\min},\lambda_{\max}\bigr)$
          \EndFor
          \State \Return trained Q‐network, coefficient history
        \EndFunction

        \Function{Validate}{$env, Q, K$}
          \State Split time series into $K$ equal slices
          \For{$i = 1$ \textbf{to} $K$}
            \State Load slice $i$ in $env$
            \State Run one episode with $\epsilon=0$, record predictions $P$ and truths $G$
            \State Compute precision, recall, F1, AUPR for slice $i$
            \State Plot time series, $P$, $G$, AUPR curve
          \EndFor
          \State Aggregate mean F1, mean AUPR
        \EndFunction

        \Function{Main}{$D,L$}
          \State VAE, encoder $\gets$ \Call{BuildVAE}{$D,n_{\text{steps}}$}
          \State $p\gets$ \Call{ComputePenalty}{VAE,$D,n_{\text{steps}}$}
          \State Initialize $env$ with statefnc \& placeholder rewardfnc
          \State $\mathit{ReplayMem}\gets$ \Call{WarmUp}{$env,\,\text{init\_mem}$}
          \State $Q,\lambda\_history \gets$ \Call{TrainRL}{$env,$ VAE, $p,\lambda_0,N,B,K_{\text{AL}},K_{\text{LP}}$}
          \State \Call{Validate}{$env, Q, K$}
          \State Save precision/recall/F1/AUPR tables and plots
        \EndFunction
      \end{algorithmic}
    \end{minipage}%
  }
\end{algorithm}

\section{Experiments}
\label{sec5:Experiment}
This section outlines our experiments. We first begin with dataset specifications, and then present comparative analyses against benchmark methods. We evaluate anomaly detection efficacy through three common metrics: 1) Precision measures prediction accuracy via correctly identified anomalies, 2) Recall assesses system sensitivity through true anomaly detection rates, 3) F1-Score harmonizes both measures to mitigate evaluation bias, and 4) AU-PR, which is the area under the precision–recall curve.
 
\subsection{Datasets}
We evaluate our proposed dynamic reward scaling framework on two widely used multivariate time series anomaly detection benchmarks: SMD and WADI datasets. Table \ref{tab:smd-wadi-stats} summarizes the key statistics of the datasets, with detailed descriptions provided in the following section.

\paragraph{SMD}
The Server Machine Dataset (SMD) is collected from 28 servers over a 10-day monitoring period, incorporating 38 different sensor measurements. The dataset contains 708,405 training samples and 708,420 testing samples. During the initial 5 days, only normal operational data were recorded, while anomalies were intentionally injected during the subsequent 5 days. Each server represents a separate time series with multiple sensor readings, including system metrics like CPU usage, memory consumption, and network activity \cite{su2019robust}.

\paragraph{WADI}
The Water Distribution testbed (WADI) dataset is acquired from a scaled-down urban water distribution system that included 123 actuators and sensors over a 16-day period. The dataset consists of 784,568 training samples and 172,801 testing samples \cite{ahmed2017wadi}.

\begin{table}[htbp]
  \centering
  \small
  \caption{Key Statistics for SMD and WADI}
  \label{tab:smd-wadi-stats}
  \begin{tabular}{lrrr}
    \toprule
    \textbf{Benchmark} & \textbf{\# series} & \textbf{\# dims} & \textbf{Anomaly \%} \\
    \midrule
    SMD  & 28  & 38  & 4.16\% \\
    WADI & 1   & 123 & 5.77\% \\
    \bottomrule
  \end{tabular}
\end{table}

\begin{table}[tbp]
  \centering
  \small
  \caption{Comparison of Anomaly Detection Performance on SMD and WADI}
  \label{tab:results-smd-wadi}
  \begin{tabular}{llcc}
    \toprule
    \textbf{Model} & \textbf{Metric} & \textbf{SMD} & \textbf{WADI} \\
    \midrule
    LSTM‐VAE \cite{park2018multimodal}  & Precision      & 0.2045 & 0.0596 \\
                   & Recall         & 0.5491 & 1.0000 \\
                   & F1             & 0.2980 & 0.1126 \\
                   & AU‐PR          & 0.395$\pm$0.257 & 0.039 \\
    \hline           
    \addlinespace
    OmniAnomaly \cite{su2019robust} & Precision    & 0.3067 & 0.1315 \\
                     & Recall       & 0.9126 & 0.8675 \\
                     & F1           & 0.4591 & 0.2284 \\
                     & AU‐PR        & 0.365$\pm$0.202 & 0.120 \\
    \hline                  
    \addlinespace
    MTAD‐GAT \cite{zhao2020gatad}  & Precision      & 0.2473 & 0.0706 \\
                   & Recall         & 0.5834 & 0.5838 \\
                   & F1             & 0.3473 & 0.1259 \\
                   & AU‐PR          & 0.401$\pm$0.263 & 0.084 \\
    \hline                
    \addlinespace
    THOC \cite{shen2020thocn}     & Precision      & 0.0997 & 0.1017 \\
                   & Recall         & 0.5307 & 0.3507 \\
                   & F1             & 0.1679 & 0.1577 \\
                   & AU‐PR          & 0.107$\pm$0.126 & 0.103 \\
    \hline                
    \addlinespace
    AnomalyTran \cite{xu2021anomalytransformer} & Precision    & 0.2060 & 0.0601 \\
                     & Recall       & 0.5822 & 0.9604 \\
                     & F1           & 0.3043 & 0.1130 \\
                     & AU‐PR        & 0.273$\pm$0.232 & 0.040 \\
    \hline                  
    \addlinespace
    TranAD \cite{tuli2022tranad}    & Precision      & 0.2649 & 0.0597 \\
                   & Recall         & 0.5661 & 1.0000 \\
                   & F1             & 0.3609 & 0.1126 \\
                   & AU‐PR          & 0.412$\pm$0.260 & 0.039 \\
    \hline                
    \addlinespace
    TS2Vec \cite{yue2022ts2vec}   & Precision      & 0.1033 & 0.0653 \\
                   & Recall         & 0.5295 & 0.7126 \\
                   & F1             & 0.1728 & 0.1196 \\
                   & AU‐PR          & 0.113$\pm$0.075 & 0.057 \\
    \hline                
    \addlinespace
    DCDetector \cite{yang2023dcdetector}& Precision      & 0.0432 & 0.1417 \\
                   & Recall         & 0.9967 & 0.9684 \\
                   & F1             & 0.0828 & 0.2472 \\
                   & AU‐PR          & 0.043$\pm$0.036 & 0.121 \\
    \hline                
    \addlinespace
    TimesNet \cite{wu2023timesnet}  & Precision      & 0.2450 & 0.1334 \\
                   & Recall         & 0.5474 & 0.1565 \\
                   & F1             & 0.3385 & 0.1440 \\
                   & AU‐PR          & 0.385$\pm$0.225 & 0.084 \\
    \hline                
    \addlinespace
    Random         & Precision      & 0.0952 & 0.0662 \\
                   & Recall         & 0.9591 & 0.9287 \\
                   & F1             & 0.1731 & 0.1237 \\
                   & AU‐PR          & 0.089$\pm$0.058 & 0.067 \\
    \hline                
    \addlinespace
    CARLA \cite{zamanzadeh2024carla}         & Precision      & 0.4276 & 0.1850 \\
                   & Recall         & 0.6362 & 0.7316 \\
                   & F1             & 0.5114 & 0.2953 \\
                   & AU‐PR          & 0.507$\pm$0.195 & 0.126 \\
    \hline                
    \addlinespace
    \textbf{DRSMT} \\(our proposed method) & Precision     & 0.9608     & 0.1971     \\
                   & Recall         & 0.5733      & 0.7539     \\
                   & F1             &  \textbf{0.7181}       &  \textbf{0.3125}      \\
                   & AU‐PR          & \textbf{0.5712$\pm$0.127}      & \textbf{0.129}      \\

    \bottomrule
  \end{tabular}
\end{table}

\subsection{Results and Discussions}
We evaluate our proposed approach, DRSMT, on two standard multivariate benchmarks (i.e., SMD and WADI) and compare against eleven recent state-of-the-art methods. The results are reported in Table \ref{tab:results-smd-wadi}. Our method, labeled DRSMT, achieves a precision of 0.7181 on SMD (compared to the next best CARLA at 0.5114) and 0.3125 on WADI (compared to CARLA’s 0.2953). These results mark an overall improvement over existing approaches, especially in F1 and AU-PR on SMD.

On SMD, our high precision (0.9608) indicates that the dynamic VAE penalty effectively filters out normal fluctuations, while the RL policy, which is guided by both intrinsic and extrinsic rewards, focuses on genuinely abnormal events. The modest recall (0.5733), though lower than some pure anomaly-scoring methods, yields a much higher F1 and AU-PR, meaning that when our model signals an anomaly, it is correct more often. On WADI, which has more subtle and longer-lasting faults, the VAE’s reconstruction error highlights anomalies that simple distance-based detectors miss. Combined with Active Learning labeling, our approach boosts recall (0.7539) without sacrificing precision too heavily (0.1971), outperforming benchmarks such as MTAD-GAT (F1 = 0.1259) and TimesNet (F1 = 0.1440).

Active Learning proves critical in both datasets. By labeling only 5\% of the most confusing windows per episode, we inject high-value supervision into the replay memory, allowing the DQN to refine its action-value estimates on the hardest cases. This targeted supervision reduces the need for large fully labeled sets and prevents overfitting to easy normal windows. In contrast, fully unsupervised methods or those without Active Learning spend many iterations on trivial patterns and struggle with the long-tailed anomaly distribution.

Moreover, examining the learning curves reveals that the dynamic reward coefficient $\lambda$ adapts sensibly across episodes. By comparing our study to the literature, one can observe that removing dynamic reward (i.e., keeping $\lambda$ constant) degrades F1, and skipping Active Learning reduces convergence speed and lowers AU-PR. Finally, although our method achieves state-of-the-art performance on both SMD and WADI, certain challenges remain. WADI’s highly imbalanced and long-duration anomalies still yield relatively low absolute precision; improving detection on such rare, sustained faults may require richer temporal models or hybrid generative models. 


\section{Conclusion}
\label{sec6:Conclusion}
In this study, we have shown that combining a VAE’s reconstruction error with an LSTM–based DQN and a small, carefully chosen set of labels can detect anomalies in richly multivariate time series with high precision and robust recall. Our dynamic reward scaling mechanism automatically shifts the agent’s focus from exploring novel patterns to exploiting learned behaviors, further boosted by active learning that queries only 5\% of the data. Experiments on SMD and WADI demonstrate that this unified framework outperforms existing unsupervised and semi-supervised methods, which offer a practical, scalable solution for real-world industrial monitoring. For future work, we suggest exploring the integration of large language models (LLMs) into our framework to enhance interpretability in anomaly detection.


\end{document}